\documentclass{nejsds}

\usepackage{stmaryrd}
\usepackage{float}
\usepackage{movie15}
\usepackage{amsfonts}
\usepackage{amssymb, bm}
\usepackage{amsmath}
\usepackage{marvosym}
\usepackage{verbatim}
\usepackage{asymptote}
\usepackage{physics}
\usepackage[ruled,vlined]{algorithm2e}
\usepackage{multirow}

\begin{document}

\begin{frontmatter}


\title{Conformal prediction for text infilling and part-of-speech prediction}

\vspace{.25in}
Neil Dey$^{1}$, Jing Ding$^{1}$, Jack Ferrell$^{2}$, Carolina Kapper$^{3}$, Maxwell Lovig$^{4}$, Emiliano Planchon$^{1}$, and Jonathan P Williams$^{1}$ \\[.15in]
$^{1}$ North Carolina State University\\
$^{2}$ University of Florida \\
$^{3}$ High Point University \\
$^{4}$ University of Louisiana, Lafayette \\



\begin{abstract}
Modern machine learning algorithms are capable of providing remarkably accurate point-predictions; however, questions remain about their statistical reliability. Unlike conventional machine learning methods, conformal prediction algorithms return confidence sets (i.e., set-valued predictions) that correspond to a given significance level. Moreover, these confidence sets are valid in the sense that they guarantee finite sample control over type 1 error probabilities, allowing the practitioner to choose an acceptable error rate. In our paper, we propose inductive conformal prediction (ICP) algorithms for the tasks of text infilling and part-of-speech (POS) prediction for natural language data. We construct new conformal prediction-enhanced bidirectional encoder representations from transformers (BERT) and bidirectional long short-term memory (BiLSTM) algorithms for POS tagging and a new conformal prediction-enhanced BERT algorithm for text infilling. We analyze the performance of the algorithms in simulations using the Brown Corpus, which contains over 57,000 sentences. Our results demonstrate that the ICP algorithms are able to produce valid set-valued predictions that are small enough to be applicable in real-world applications. We also provide a real data example for how our proposed set-valued predictions can improve machine generated audio transcriptions.\\

\end{abstract}

\end{frontmatter}

\section{Introduction}

In recent years, machine learning algorithms have dominated the realm of natural language processing (NLP). Over time, these algorithms have achieved higher and higher accuracy in various NLP tasks. However, such algorithms are specialized for point prediction, and as such, a significant limitation of many machine learning algorithms is that they do not offer any uncertainty quantification about how often these point predictions are actually correct. To address this limitation, the ideas of conformal prediction have gained traction in recent years in the machine learning literature generally, but less so in application to NLP tasks. 

Conformal prediction is an approach introduced in \cite{Vovok1999} that allows, for example, a point prediction method to be extended to form confidence sets, guaranteeing that the set contains the true unknown predictor value with some nominal coverage probability. It has been shown that deep learning architectures such as multilayer perceptrons (MLP), convolutional neural networks (CNN), and gated recurrent units (GRU) often improve in their robustness when enhanced by a conformal prediction algorithm \citep{Messoudi2020}. Conformal prediction has been applied to text classification NLP tasks. For example, \cite{Maltoudoglou2020} and \cite{Maltoudoglou2022} demonstrate similar results for conformal prediction-enhanced BERT and artificial neural network (ANN)-based sentiment classification and multi-label text classification, respectively. Other experiments in the literature, such as \cite{Paisios2019} working with deep neural network (DNN)-based multi-label text classifiers and \cite{Cauchois2021} working with tree-based classifiers, replicate these findings for other multi-label classification models. Conformal prediction has also been successful in relation classification, identifying relationships between two entities in a sentence, as demonstrated by \cite{Fisch2021a} and for open-domain question answering and information retrieval for fact verification in \cite{Fisch2020}. To our knowledge, however, conformal prediction has not been applied to two key tasks in NLP: the text infilling task and POS tagging.

The text infilling task (also known as the Cloze task) is a standard NLP task, asking a model to ``fill in the blank'' given an otherwise complete sentence. Since its conception, the task has greatly expanded in scope due to the great success of various text infilling algorithms developed. For example, \cite{Fedus2018} uses generative adversarial networks to great effect in the MaskGAN algorithm to generalize the problem to full text generation. Another generalization of the text infilling task was introduced by \cite{Mostafazadeh2016} in the form of the Story Cloze Test, determining the ``right ending” to a story. The Story Cloze Test has been further explored in the form of neural network solutions \citep{Srinivasan2018} and generative pre-training of language models \citep{Radford2018}, among other methods. Yet another extension to the text infilling task comes in the form of filling in blanks of arbitrary length, as explored in \cite{Zhu2019} (utilizing self-attention mechanisms) and in \cite{Shen2020} (using the blank language model). Although many techniques have been proposed to solve the text infilling task, such as gradient-search-based inference \citep{Liu2019} and infilling by language modeling \citep{Donahue2020}, text infilling in practice has been dominated by the BERT algorithm \citep{Devlin2019}, which uses a masked language modeling (MLM) pre-training objective to attain word embeddings. Though trained on the text infilling task, the resulting word embeddings remain competitive in many standard NLP tasks. 

The POS tagging task is another standard NLP task in which a model assigns the correct grammatical POS to each word in a sentence. This task is unusual in the NLP realm in that the most naive algorithm of simply assigning each word its most common POS already achieves a very high baseline accuracy of roughly 92\% \citep[][Chapter 8, end of Section 2]{Jurafsky2021}. The introduction of some classical models such as hidden Markov models (HMM) \citep{Kupiec1992} and conditional random fields (CRF) \citep{Lafferty2001} improved the accuracy to about 96\%; more modern techniques currently used such as the BiLSTM proposed by \cite{Wang2015} and transformer models such as BERT \citep{Devlin2019} offer further marginal improvements, reaching about 97-98\% accuracy. Similar to the text infilling task, it does not appear that the application of conformal prediction to POS tagging is present in the literature. However, a method of set-valued prediction introduced by \cite{Mortier2019} has been applied to POS tagging of a middle-lower German corpus by \cite{Heid2020}, demonstrating more robust predictions than standard POS tagging algorithms, but these set-valued predictions do not offer the guaranteed control over type 1 error probabilities that are inherent in conformal prediction sets.  As discussed in \cite{Heid2020}, POS tagging of historical corpora remains one area where linguistics experts do not necessarily know or agree on the POS for particular words because the languages are no longer in use.  In these applications, set-valued predictions are most sensible.  

Furthermore, in machine learning applications, since the accuracy of POS tagging is typically high, it can be expected that many set-valued POS predictions will be of size 1, and greater than 1 for occasional ambiguous cases.  Accordingly, the set-valued POS tagging algorithms that we contribute combine the speed of automated tagging with the accuracy of manual tagging.

In our paper, we apply conformal prediction to the text infilling (more specifically the MLM) and POS tagging tasks.  We construct new conformal prediction-enhanced BERT and BiLSTM algorithms for POS tagging and a new conformal prediction-enhanced BERT algorithm for MLM. Using the Brown Corpus \citep{Francis1979}, we empirically demonstrate that BERT provides smaller prediction sets for POS tagging than a BiLSTM model, and we show that BERT generates usefully small prediction sets for MLM.  Moreover, we show that all conformal prediction sets achieve their nominal coverage for any level of significance.  A brief overview of BiLSTM models, transformers and BERT, and conformal predictions is given in Section \ref{background}.  Section \ref{methods} presents our proposed algorithms, followed by a discussion of our empirical studies in Section \ref{results}. The utility of the enhanced BERT model for MLM in a realistic setting is illustrated in Section \ref{real_data} by running the model on missing words from a transcript of a TED Talk generated by automatic speech recognition software, and the paper closes with concluding remarks provided in Section \ref{conclusion}. The code and workflow for reproducing our results, along with documented software for implementing our algorithms on new data sets, are available at {\footnotesize \verb1https://github.com/jackferrellncsu/drums-nlp-codesnapshot1}.

\section{Existing machine learning approaches}\label{background} 

Currently, the state-of-art methods for MLM tasks are BERT-based \citep{Devlin2018}. Other models include TagLM \citep{Peters2017} and ELMo \citep{Peters2018}. TagLM and ELMo both use recurrent neural networks (RNN), and ELMo specifically constructs a two-layer BiLSTM, commonly used as a pre-trained model for the embedding layer for other models. Alternatively, BERT models use transformers instead of an LSTM in the deep embedding layer.

POS tagging takes a sequence of words and assigns each word a particular POS. It is a sequence labeling task because each word can represent different a POS depending on its context. POS tagging is useful in syntactic parsing, reordering in translation, sentiment tasks, text-to-speech tasks, etc. Classic POS labeling algorithms include HMM and linear chain CRF. HMM is a probabilistic sequence model that computes a probability distribution over possible sequences of labels and chooses the label sequence with highest likelihood. However, as a generative model, HMM does not incorporate arbitrary features for unknown words in a clean way. \cite{Brants2000} implemented HMM, handling unknown words using suffix features, and attained an accuracy of 96.46\%. CRF is a log-linear model that assigns a probability to an entire output (label) sequence with respect to all the possible sequences, given the entire sequence of input words. \cite{Sun2014} proposed using CRF with a method for structure regularization and achieved 97.36\% accuracy. 

Modern POS labeling algorithms include RNNs and transformer networks. Both approaches manage to deal directly with the sequential nature of language without being restricted to a fixed window size surrounding the target word. RNN architectures contain a cycle within the network connections, where the value of a unit is directly or indirectly dependent on the earlier output as an input. The BiLSTM architecture has achieved wide attention due to its effectiveness for sequence classification. It solves the ``vanishing gradient'' problem by forgetting information that is no longer needed, carrying information that is required for decisions to come, and combining the forward and backward network results. Researchers have applied BiLSTMs and obtained accuracies ranging from 97.22\% to 97.76\% \citep{Ling2015, Plank2016, Yasunaga2018, Bohnet2018, Xin2018, Liu2018}. As an alternative solution, transformers are made up of blocks including self attention layers, feedforward networks, and custom connections. Transformer based models, such as BERT, are pre-trained on large context corpora and are well-suited for POS tagging.

Although it appears promising that the accuracy of POS tagging has reached 97\% for English language texts, it should be noted that the baseline accuracy is 92\% \citep[][Chapter 8, end of Section 2]{Jurafsky2021} because many words have only a single POS, and those that have multiple POS overwhelmingly occur with their most common class. However, a single bad tagging in a sentence can lead to a huge error in downstream tasks such as dependency parsing. It is thus more meaningful to view the accuracy of the whole-sentence POS tagging, which is around 55-57\% \citep{Manning2011}.  Researchers have been trying to improve the accuracy of POS tagging via improvements in features, parameters, and learning methods without breakthrough success. Meanwhile, there are concerns regarding the correctness of the treebank and whether POS labels are well-defined to allow us to assign each word a single symbolic label \citep{Manning2011}. That is to say, it is possible that the error in POS labeling is due to linguistically justified definitions and cannot be further improved without improvement in the field of linguistics.

One way to deal with the current error in POS tagging for further improvement is to add associated confidence values for each prediction. All the aforementioned approaches only output a simple point prediction without evaluating how likely it is for each prediction to be correct. The likelihood of each prediction enables us to evaluate how much we can rely on the prediction and generates alternative POS tags. This serves as a filtering mechanism with regard to the corresponding confidence level and can help avoid the problem that a single mistake in a sentence limits the usefulness of a tagger for downstream tasks. Conformal prediction \citep{Shafer2008} is well-suited to provide such confidence information on top of the traditional algorithms. Moreover, \cite{Papadopoulos2008} introduced the more computationally feasible application of ICP in neural networks. \cite{Maltoudoglou2020} applied ICP on a binary text classification problem using a BERT model for contextualized word embeddings. The results show that the prediction accuracy for the BERT classifier was maintained, while the prediction sets calculated using the conformal prediction algorithm provided more useful information. \cite{Fisch2020} expanded the conformal prediction correctness criterion by adding admissible labels to reduce the size of predicted sets, and filtered out implausible labels early on by using conformal prediction cascades to decrease the computational cost. \cite{Maltoudoglou2022} continued the study of conformal prediction applied to ``multi-label'' text classification using DNNs based on contextualized and non-contextualized word embeddings. They reduced the computational complexity by eliminating label-sets that would surely have p-values below the specified significance level. Their results show that the context-based classifier with conformal predictions has good performance and small prediction sets that are practically useful.

\subsection{Long short-term memory neural net}

The use of RNNs in NLP tasks is very common due to the sequential nature of language. Unlike feed-forward networks, RNNs are able to take into account all of the preceding words in a variable length sequence with fixed-size input and embedding vectors when making predictions \citep{Elman1990}. In language tasks like next word prediction, this is desirable because the more structured the context that a model is learning from, the more accurate the prediction is likely to be. 

In machine learning, the goal of a gradient descent algorithm is to minimize the cost function by finding and updating the parameters of the model. With RNNs, using gradient descent with an error criterion for tasks involving long-term dependencies is inadequate and may result in exploding or vanishing gradients \citep{Bengio1994}. This problem arises when the network updates the weights while back-propagating through time during training \citep{Hochreiter1998}. An extremely large gradient will make the model that is being trained unstable, and an extremely small ($\approx 0$) gradient will make it impossible for the model to learn correlations between events with a high temporal span of dependencies \citep{Pascanu2012}. Moreover, gradient descent becomes less efficient the further apart the inputs are, suggesting that RNNs are not desirable for tasks that require long-term “memory.” There have been many theorized solutions to these issues; however, none are as prevalent as gated neural networks \citep{Hu2018}. 

\begin{figure}[H]
    \centering
    \includegraphics[scale = 0.41]{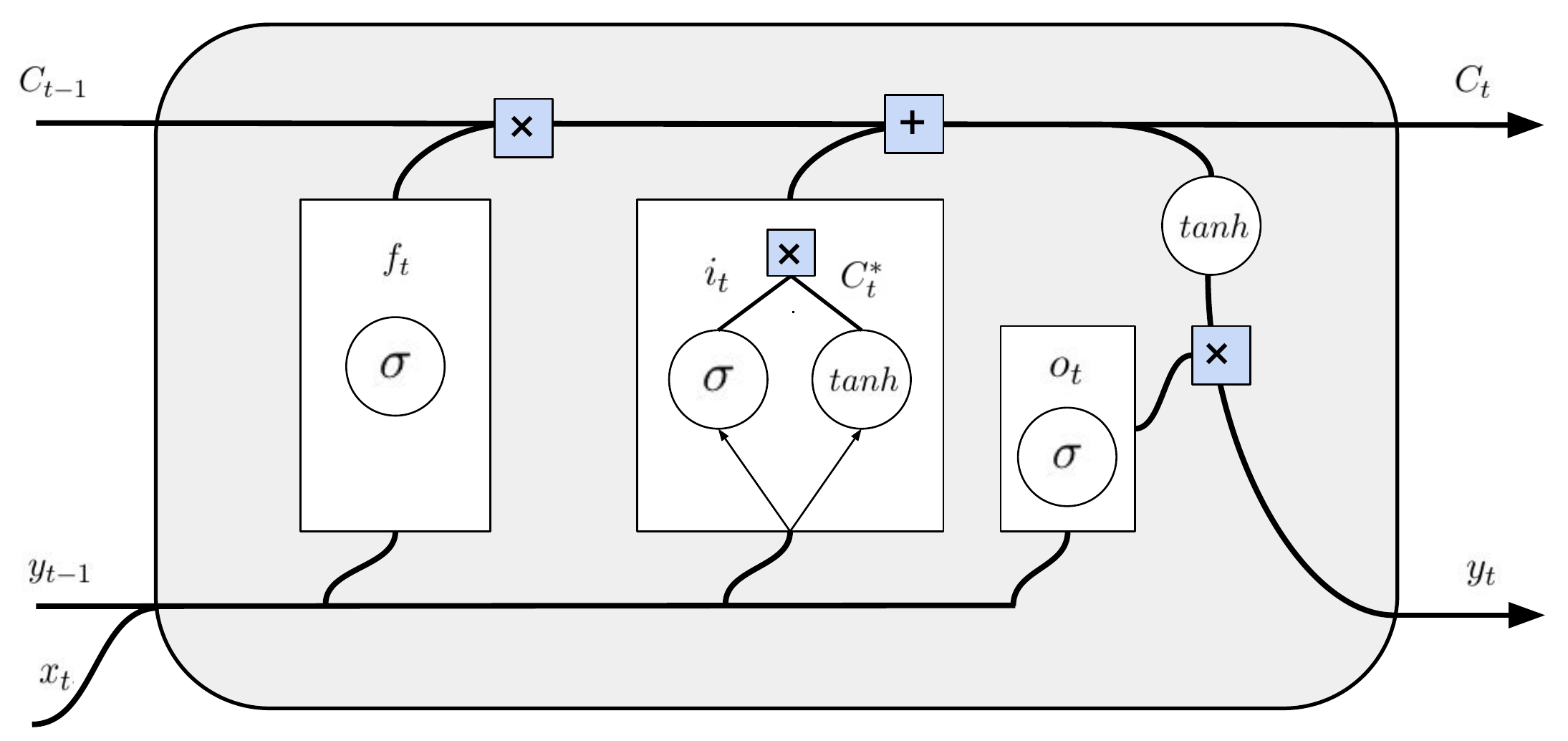}
    \caption{\footnotesize LSTM Memory Cell}
    \label{fig:memory_cell}
\end{figure}

A popular type of gated neural network is the LSTM \citep{Hochreiter1997}. LSTMs help prevent vanishing and exploding gradients through the use of a memory cell, which is regulated by the forget ($f_{t}$), input ($i_{t}$), and output ($o_{t}$) gates (see Figure \ref{fig:memory_cell}). Each of these gates contain a sigmoid activation alongside a component-wise multiplication operation. The sigmoid layer outputs values that are between 0 and 1 which serve as indicators for the proportion of each component that will be ``let through'' the gate. The standard reference for describing the architecture and intuition for memory cells is given in \cite{Olah2015}. For convenience, we summarize the main ideas in the remainder of this section. 

The forget gate ($f_{t}$) considers $y_{t-1}$ and $x_{t}$, where $y_{t-1}$ is the network output layer at time $t-1$ and $x_{t}$ is the input vector at time $t \in \mathbb{N}$. These quantities are passed through the vectorized sigmoid function
\[
 f_{t} = \sigma( [x_{t}^{T}, y_{t-1}^{T}] \cdot W_{f} + b_{f}),
\]
where $W_{f}$ and $b_{f}$ are a weight matrix and bias vector, respectively. After passing $[x_{t}^{T}, y_{t-1}^{T}]$ through the forget gate, the past cell state $C_{t-1}$ is multiplied component-wise with $f_{t}$.  Next, as shown in Figure \ref{fig:memory_cell}, a $\tanh$ activation function also evaluated at $[x_{t}^{T}, y_{t-1}^{T}]$, but with a different weight matrix $W_C$ and bias vector $b_C$, is used to create a vector of values in $[ -1,1 ]$:
\[
C^{*}_{t} = \tanh( [x_{t}^{T}, y_{t-1}^{T}] \cdot W_{C} + b_{C}).
\]
The input gate is similarly constructed as
\[
i_{t} = \sigma( [x_{t}^{T}, y_{t-1}^{T}] \cdot W_{i} + b_{i})
\]
for weight and bias terms $W_{i}$ and $b_{i}$ and the cell is updated as
\[
C_{t} = f_{t} \odot C_{t-1} + i_{t} \odot C^{*}_{t},
\]
where $\odot$ denotes component-wise multiplication.  The $f_{t} \odot C_{t-1}$ term controls how much of the past cell memory to carry forward, and the $i_{t} \odot C^{*}_{t}$ term controls how much of the updated cell memory to add \citep{Goldberg2016}. Lastly, the cell updates the state $y_{t}$ as
\begin{align*}
o_{t} & = \sigma( [x_{t}^{T}, y_{t-1}^{T}] \cdot W_{o} + b_{o}) \\
y_{t} & = o_{t} \odot \tanh(C_{t}),
\end{align*}
using a final set of weight and bias terms, $W_{o}$ and $b_{o}$.

The implementation of a memory cell like the one above is quite common; however, there is much variety when it comes to the exact details \citep{Olah2015}. Examples include GRUs \citep{Cho2014}, peephole connections \citep{Gers2002}, and clockwork RNNs \citep{Koutnik2014}, among others. The sophisticated nature of these memory cells have proven to work efficiently on NLP problems \citep{Sharfuddin2018}, which is why we consider it a favorable method to combine with a conformal predictor.

\subsection{Transformers and BERT embeddings}

Recurrent models, while useful for encapsulating information about the structure of sentences, are extremely computationally expensive in practice. Namely, the sequential nature of such models makes training them impossible to parallelize.  Transformers were introduced to fix this issue with an encoder/decoder structure \citep{Vaswani2017}. To understand the encoder/decoder intuitively, consider the problem of machine translation.  If we have a sentence in written in Spanish, the encoder will attempt to construct a mathematical representation for the meaning of the sentence. The decoder will take this mathematical representation, as well as information about the English language (for example), and combine the two to create an English sentence. The meaning of the sentence and information about the English language are captured using a technique referred to as ``attention'' \citep{Vaswani2017}. The following description of attention closely follows the source paper \cite{Vaswani2017}, and is provided for convenience.

In attention, an output is computed using a weighted sum of values, but with weights learned from a function that finds the compatibility between a query and the key corresponding to a value, where the query, the key-value pairs, and the output are all represented by vectors \citep{Vaswani2017}. Attention is mathematically described as
\[
\operatorname{Attention}(K, V, Q) = \operatorname{softmax}\qty(\frac{QK^\top}{\sqrt{d_k}})V,
\]
where $K$ is the matrix containing the key vectors with $d_k$ number of rows, $V$ is the matrix containing the value vectors, and $Q$ is the matrix containing the query vectors \citep{Vaswani2017}. The scalar $d_k$ is introduced as a normalization factor, lest dot-products become so large as to be unusable \citep{Vaswani2017}. Different items are used as keys, values, and queries depending on the context. In the most basic case, the query is the word currently being examined, the key vector is all words being used as context for the query word, and the value vector is also all the words being used as the context for the query word.  The output of the softmax function in the above equation is used as a weighting matrix for the value vectors comprising $V$.

It is often desirable for different weights to be learned based on some number, $h$, of different features of text, so the notion of ``multi-head'' attention is defined as
\[
\operatorname{MultiHead}(Q, K, V) = [\operatorname{head}_1, \dots, \operatorname{head}_h]\cdot W^O,
\]
where for each $j \in \{1,\dots,h\}$,
\[
\operatorname{head}_j := \operatorname{Attention}(Q\cdot W_j^Q, K\cdot W_j^K, V\cdot W_j^V),
\]
with weight matrices $W^O, W_j^Q, W_j^K$, and $W_j^V$ to be learned. Each head is empirically constructed to focus on different aspects of the the training text \citep{Vaswani2017}.  These multi-head layers are stacked and then fed into a feed-forward neural network to form the encoder and decoder.

Transformers have been used in many state-of-the-art NLP models, such as GPT \citep{Radford2018}, BERT \citep{Devlin2019}, and ERNIE \citep{Zhang2019}.  In developing our conformal predictors, we choose to incorporate pre-trained word embeddings from BERT in particular. We focus on the use of ``BERT-base'' rather than ``BERT-large'' due to the high computation cost associated with the latter. Nonetheless, both deliver state-of-the-art results, so any minor trade-off in accuracy is justified.  BERT-base has 12 layers, 768 hidden states, and 12 self-attention heads for a total parameter count of 110 million \citep{Devlin2019}.

The main difference between BERT and the original transformer is its ability to examine context in both directions simultaneously, whereas the original transformer \citep{Vaswani2017} and GPT \citep{Radford2018} both gated the decoder layer, only allowing it to look in the direction from which it was supposed to be predicting. This proved effective, giving both versions of the original BERT state-of-the-art results across all generalized language understanding evaluation (GLUE) \citep{Wang2019} tasks when the paper was published in 2019 \citep{Devlin2019}.  BERT was pre-trained using two tasks, next sentence prediction (NSP) and MLM.  In NSP, BERT is presented with two sentences and attempts to determine whether or not they are truly sequential. In MLM, BERT is presented with a masked word and asked to predict it given a context.  During pre-training, 15\% of words were masked so as to not let the model look at the correct answer while predicting.  BERT was trained over the entirety of Wikipedia (approximately 2.5 billion words) and the BooksCorpus \citep{Zhu2015} in efforts to mimic language as closely as possible.  A new sub-field, ``BERTology'', has surfaced in an attempt to explain why the embeddings are so efficient and generalizable \citep{Rogers2020}.  We hope our application of conformal predictors to the BERT MLM task will contribute to this area of study.

\subsection{Conformal predictions}

Throughout the remainder of the paper we will use the following notation. Let $D$ denote a corpus of text, where  the index $i \in \{1, \dots, n\}$ denotes the position of the $i$-th word and $n$ denotes the total number of words in $D$. For the POS tasks, let $y_i^{\text{pos}}$ represent the true POS for the $i$-th word in $D$. Similarly, for the MLM tasks, let $y_j^{\text{mlm}}$ represent the true masked word for the $j$-th sentence in $D$, for $j \in \{1, \dots, k\}$ where $k$ is the total number of sentences in $D$. For training, testing, and calibration, the entire corpus $D$ is randomly split into three pieces $D_{\text{train}}$,  $D_{\text{test}}$, and $D_{\text{cal}}$, respectively.

Point predictions have been the standard for NLP tasks, including those from neural networks and transformers. However, uncertainty quantification in the form of confidence intervals/sets provide added utility for point predictions for NLP tasks. Recent work has shown that this can be achieved in a variety of NLP tasks such as sentiment classification \citep{Maltoudoglou2020}, multi-label text classification \citep{Maltoudoglou2022}, open-domain question answering \citep{Fisch2020}, and information retrieval for fact verification \citep{Fisch2020}. \cite{Maltoudoglou2020} used conformal prediction for document-level binary sentiment analysis to determine whether IMDB movie reviews had positive or negative connotation. Their data set contained 25,000 positive and 25,000 negative reviews, and they used ICP with a BERT-based text classification model to create confidence sets. At roughly 91\% confidence (i.e., $\epsilon$ = 0.09), the average set size had been narrowed to one classification, and both the sigmoid and softmax activation functions were found to perform equally well \citep{Maltoudoglou2020}. We construct similar algorithms for multi-label classification for POS tagging and MLM. \cite{Fisch2021} expanded the use of conformal predictions for information retrieval with a cascading approach, filtering out incorrect options at every step with the hopes of keeping at least one ``admissible'' option after all the layers. This approach was found to improve both computational and predictive efficiency by giving the model fewer items to sort through at each step \cite{Fisch2021}.

Conformal prediction uses knowledge gained from training a model to create confidence sets with guaranteed finite sample control over the probability of a type 1 error \citep{Shafer2008} and can be built on almost any machine learning tool, including neural networks \citep{Vovk2005}.  Precisely, assuming exchangeable data examples, for any level of significance $1 - \epsilon$ with $\epsilon \in (0,1)$, a conformal predictor yields a set-valued prediction with the property that it will fail to include the true label with probability at most $\epsilon$ \citep{Shafer2008}.  This property, referred to as ``validity'', is mathematically guaranteed to hold for any finite sample size, but it is possible that the conformal prediction set is very large. The values included in the prediction sets are based on the ``strangeness'' of the test data when compared to training data, and the efficiency (i.e., size of the prediction sets) is dependent on how the strangeness measure -- a so-called ``nonconformity function'' -- is defined \citep{Vovk2005}. 

The only necessary assumption for the validity of conformal prediction sets is that the data must be exchangeable: a more relaxed assumption than the common assumption of independent and identically distributed, essentially meaning that for observed data examples $z_1, \dots, z_n$, each of the $n!$ possible orderings of the values were equally probable for being observed \citep{Shafer2008}. In that case, the collection of observed examples are best described by a ``bag''
\[
B := \lbag z_1, \dots, z_n\rbag,
\]
denoting a set of values such that the order of the elements is irrelevant \citep{Vovk2005}.  For example $\lbag 1,2\rbag = \lbag 2,1\rbag$.

A nonconformity measure $A$ is a real-valued function that measures how strange or different a value $z$ is from the other examples in the bag $B$. For the example values $z_{i} \in B$ for $i \in \{1,\dots,n\}$, denote the nonconformity scores by
\begin{equation}\label{nonconfomity_score}
\alpha_i := A(B \backslash \{z_i\}, z_i).
\end{equation}
The particular form of $A$ is context/application-specific, but common choices include various norms, such as the $\ell_{\infty}$ norm in \cite{Maltoudoglou2020} or the $\ell_{2}$ norm \citep{Shafer2008}, of distances from a `center' of the set $B \backslash \{z_i\}$ to the point $z_i$.

Next, to decide whether to include a test value $z$ in the conformal prediction set $\Gamma^{\epsilon}(z_{1},\dots,z_{n})$ with level of significance $1 - \epsilon$, first denote $z_{n+1} := z$ and update:
\[
B := \lbag z_1, \dots, z_n, z_{n+1}\rbag.
\]
Then, noting that $\alpha_{n+1}$ corresponds to the test value, include $z = z_{n+1} \in \Gamma^{\epsilon}(z_{1},\dots,z_{n})$ if
\[
p := \frac{|\{i = 1, \dots, n+1 : \alpha_{i} \geq \alpha_{n+1}\}|}{n+1} > \epsilon. 
\]
This procedure is formally described in \citep{Vovk2005,Shafer2008} as a transductive conformal algorithm, and we summarize it here as Algorithm \ref{cp_trans}.

\begin{algorithm}
\SetAlgoLined
\KwIn{Nonconformity measure A, significance level $\epsilon$, observed examples $z_1, \dots, z_{n}$, and a new observation or value $z$}
Decide whether to include $z$ in the set $\Gamma^\epsilon(z_1,\dots, z_{n})$ \\
Set $z_{n+1} := z$ \\
Set $B := \lbag z_1, \dots, z_n, z_{n+1}\rbag$ \\
\For{$i \in \{1,\dots,n+1\}$}{Set $\alpha_i := A (B \backslash \{z_i\}, z_i) $}
Set $p := \frac{|{i = 1, \dots, n+1 : \alpha_i \ge \alpha_{n+1}}|}{n+1}$\\
Include $z$ in $\Gamma^\epsilon(z_1,\dots, z_{n})$ if $p > \epsilon$
\caption{\footnotesize Transductive conformal algorithm}\label{cp_trans}
\end{algorithm}

For many machine learning applications, however, transductive conformal prediction would be too computationally expensive since it requires recomputing all of the nonconformity scores for every new test observation/value. Motivated by this issue, ICP \citep{Papadopoulos2008} is a modification of conformal prediction that greatly reduces computation costs. In ICP, the data is first split into proper training, calibration, and testing sets $D_{\text{train}}$, $D_{\text{cal}}$, and $D_{\text{test}}$, as in our notation. Next, nonconformity scores are computed for the calibration set examples analogous to equation (\ref{nonconfomity_score}) for every $j \in \{i \in \{1,\dots, n\} : d_i \in D_{\text{cal}}\}$ as
\[
\alpha_j := A(D_{\text{train}}, d_j).
\]
Without loss of generality, re-index these scores by ${j \in \{1,\dots,|D_{\text{cal}}|\}}$.  Similarly, the nonconformity score for a test observation $d^{*} \in D_{\text{test}}$ is defined as
\[
\alpha^{*} := A(D_{\text{train}}, d^{*}),
\]
and $d^{*} \in \Gamma^{\epsilon}$ if
\[
p := \frac{|\{j = 1,\dots,|D_{\text{cal}}| : \alpha_{j} \geq \alpha^{*}\}| + 1}{ |D_{\text{cal}}| +1} > \epsilon. 
\]

Thus, the ICP algorithm must only be applied once to the calibration set, and each subsequent test value only requires calculating a single new nonconformity score to compare to the static collection of nonconformity scores in the calibration set. While ICP is slightly less reliable empirically than the transductive approach, the small sacrifice in empirical reliability does not outweigh the added benefit in computational efficiency \citep{Papadopoulos2008}. From this point forward, any reference to conformal prediction should be interpreted as ICP unless otherwise stated.

\section{Methodology}\label{methods}  

In this section we present our methodological contributions, namely an ICP with a BERT-based neural network nonconformity measure for POS tagging in Algorithm \ref{alg_icp_pos}, an ICP with a BiLSTM-based neural network nonconformity measure for POS tagging in Algorithm \ref{alg_icp_pos}, and an ICP with a BERT-based neural network nonconformity measure for MLM in Algorithm \ref{alg_icp_mlm}.

\subsection{POS prediction}

POS prediction involves finding the context of a word and then outputting the corresponding POS. Here we present our ICP Algorithm \ref{alg_icp_pos} for POS prediction. Let $S$ represent the set of all $q$ unique POS in $D$, and for the $i$-th word in $D$, let $\hat{y}_i^\text{pos} \in \mathbb{R}^{q}$ represent the softmax vector produced by one of our two POS models, namely the subsequently described BERT POS (BPS) model or BiLSTM model. In addition, let $\hat{y}_{i, s}^\text{pos}$ denote the specific softmax value for any POS $s \in S$.

\begin{algorithm}
\SetAlgoLined
\KwResult{Returns the conformal prediction set $\Gamma^{\epsilon}$ containing POS labels for a test word $d^* \in D_{\text{test}}$ and significance level $\epsilon$.}
\textbf{train} the model using $D_{\text{train}}$ to produce \\
\hspace{.3in} $\{\hat{y}_i^\text{pos} : \ i \in \{1,\dots, n\} \text{ and } d_i \in D_{\text{cal}}\}$\;
\For{$j$ in $\{i \in \{1,\dots, n\} : d_i \in D_{\text{cal}}\}$}{
$s = y^{\text{pos}}_j$; \hspace{.4in} {\footnotesize \# Recall $y^{\text{pos}}_j$ is the true masked POS} \\
$\alpha_j = 1 - \hat{y}^{\text{pos}}_{j, s} $\;}
Re-index the nonconformity scores by $j \in \{1,\dots,|D_{\text{cal}}|\}$\;
\For{$s$ in $S$}{
$\alpha^*_{s} = 1 - \hat{y}^{\text{pos}}_{*, s}$\;
$p_s = \frac{|\{j = 1,\dots,|D_{\text{cal}}| : \alpha_j \geq \alpha^*_s\}| + 1}{ |D_{\text{cal}}| + 1}$\;
\If{$p_s > \epsilon$}{
$s \in \Gamma^{\epsilon}$\;}
}
\Return{$\Gamma^{\epsilon}$}
\caption{\footnotesize ICP POS Prediction}\label{alg_icp_pos}
\end{algorithm}

\subsubsection{BERT POS prediction}

BERT creates custom embeddings for words based on the words themselves and the context around them. These embeddings can be fine-tuned to specific NLP tasks, such as POS prediction. We extend these predictions to form conformal prediction sets to quantify prediction uncertainty. The parameters of BERT that we implement for POS prediction have been pre-trained and are available from \cite{Devlin2019}. However, we must adjust the BERT parameters in addition to the parameters of a dense feed-forward network that we construct for mapping the BERT-base length 768 output embedding for a word to our $q$ component softmax vector \citep{Devlin2019}.

There is some nuance to how we format the data to be usable with BERT. First, we address the BERT tokenizer. BERT splits a word root from its tense. For this reason, we define a word as its last token, since this dictates the tense of a word (e.g., the word ``wanted" is tokenized as ``\#\#ed"). Next, it is necessary for a BERT input to have a fixed length (e.g., 100 words per sentence). If a sentence is larger than this maximum length, we split the sentence into multiple sentences of length 100 and add [PAD] tokens for sentences less than 100 words until the sentence is of length 100.

\begin{figure}[H]
    \centering
    \includegraphics[trim={30 30 40 20}, clip, scale=.225]{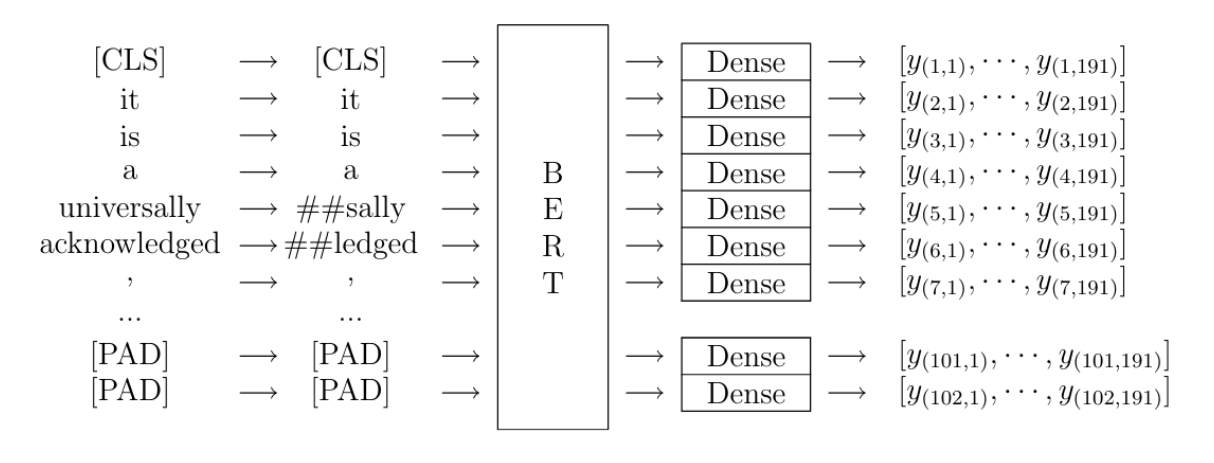}
    \caption{\footnotesize Illustration of the BERT POS model. The left most layer is the input sentence which is then transformed into the last token of each word. This 2nd layer is then input into BERT and an optimized embedding for the POS is made for each word. Each embedding is passed through a single layer dense neural net with sigmoid and softmax activation to produce the probability of each POS tag for each word in the sentence.}
    \label{fig:bert_pos_schematic}
\end{figure}

On top of BERT, we place a single softmax layer which reduces the 768 length vector into a $q$ length probability vector. Our model is trained by inputting a sentence and each word has its fine-tuned embedding vector run through the dense layer. We train the parameters for 3 epochs using the binary cross entropy loss with the RADAM optimizer \citep{Liu2020}. A schematic illustration of our BERT architecture it is given in Figure \ref{fig:bert_pos_schematic}.  The softmax output vector from this neural network is then used in Algorithm \ref{alg_icp_pos} to yield the resulting conformal prediction sets.  This combined BERT architecture with the conformal prediction algorithm for POS tagging is what we refer to as our BPS model.

\subsubsection{BiLSTM POS prediction}

In addition to our BPS model, we also construct a BiLSTM architecture for the task of POS tagging with conformal prediction sets, also using Algorithm \ref{alg_icp_pos}. For word embeddings, we use Stanford's GloVe embeddings \citep{Pennington2014}. The GloVe embeddings are desirable because of their ability to balance local and global relationships between words. To make the model more generalizable, we chose to use pre-trained embeddings. Specifically, we use the GloVe embeddings which are of length 300 and trained on 6 billion tokens from Wikipedia and Gigaword \citep{Parker2011}. Note that any word in our corpus that does not have a defined, pre-trained GloVe embedding is instead represented by a 300 length zero vector.

To train the BiLSTM model, we first create sentence embeddings to represent all of the sentences in our corpus. We create these sentence embeddings by concatenating the ordered, pre-trained GloVe word embeddings for the words in a given sentence. Accordingly, the sentence embedding for the $j$-th sentence is a matrix of dimension $300 \times n_j$, where $n_j$ is the number of words in the $j$-th sentence. These sentence embedding matrices are then passed through a layer in the BiLSTM model. The BiLSTM layer consists of two sub-layers, a forward LSTM layer and a backward LSTM layer. For any individual sentence indexed by $j$, the forward LSTM layer takes in the matrix of embeddings and returns a matrix of dimension $150 \times n_j$. Similarly, the backward LSTM layer takes in the reversed matrix of embeddings and returns a matrix of dimensions $150 \times n_j$. Each column in these returned matrices contains a 150 length embedding suited for predicting the respective POS for each word. The idea is that the forward layer is capturing the context of a sentence that is processed from beginning to end, while the backward layer is capturing the context of a sentence that is processed from end to beginning. This extra context allows for the model to get a better understanding of the sequential patterns of POS in sentences. To combine the information gathered by the forward LSTM layer and the backward LSTM layer, we reverse the order of the columns of the matrix that were returned by the backward LSTM and concatenate it with the matrix that was output by the forward LSTM. This results in a $300 \times n_j$ matrix, with each column representing an optimal embedding for predicting POS. 

\begin{figure}[H]
\centering
\includegraphics[trim={15 0 0 0}, clip, scale=0.37]{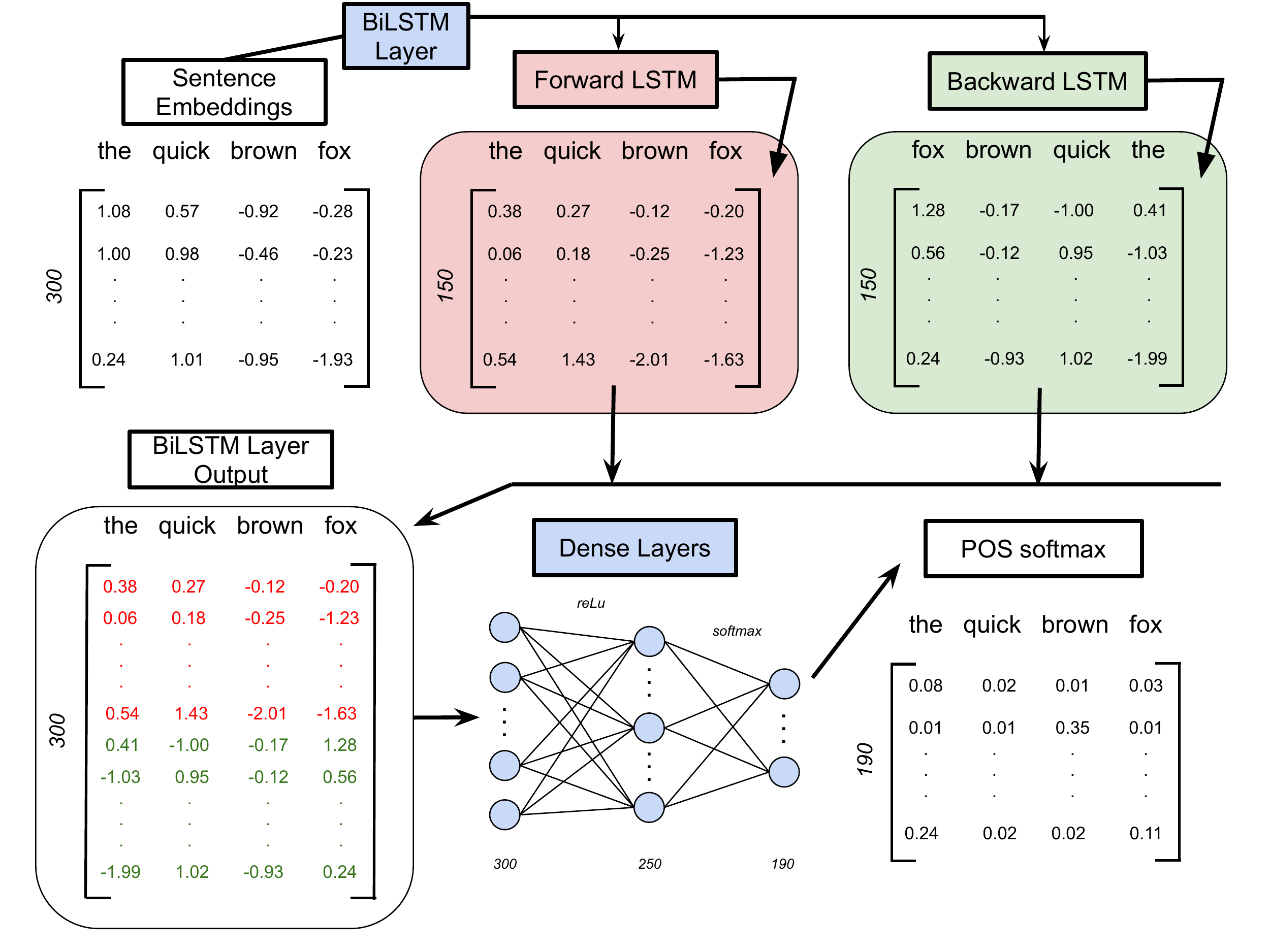}
\caption{\footnotesize Illustration of the BiLSTM POS model processing a sample sentence embedding matrix. The top row illustrates the functionality of the BiLSTM layer within the model, with the leftmost matrix in the second row symbolizing the output of the BiLSTM layer. As seen, this output is simply the concatenation of the output matrix for forward LSTM layer and the reversed output matrix for backward LSTM layer. The rest of the second row provides a visualization of the dense layer processes which eventually result in the POS softmax matrix shown in the bottom right.}
\label{fig:BiLSTM_pos_schematic}
\end{figure}

After training the BiLSTM matrix of optimal embeddings, we pass the columns of this matrix through a feed-forward neural net. This net reduces the 300 length embedding to a 250 length vector with a ReLU activation, which is further reduced to a $q$ length softmax vector corresponding to the $q$ POS labels. Each softmax output vector represents an estimated probability distribution over the POS labels for a given word. This procedure is repeated for the $n_j$ columns in the input matrix (each column corresponding to a word in the input sentence). The schematic for this BiLSTM architecture is displayed in Figure \ref{fig:BiLSTM_pos_schematic}. 

For training the parameters, we implement exponential decay in the popular RADAM optimizer \citep{Liu2020}. We train for 700 epochs to avoid overfitting and we use cross entropy as our loss function. Finally, similarly to our BPS model, the softmax output vector from this neural network is then used in Algorithm \ref{alg_icp_pos} to yield the resulting conformal prediction sets.

\subsection{Masked language modeling}

The MLM task is similar to POS tagging with two exceptions. First, the word to be predicted is masked or unknown (for training/testing, when a sentence is passed into the model, the target word is assigned the [MASK] token). Second, instead of classifying a word using $q$ POS labels, unknown words are inferred using a massive vocabulary of words. Though, these changes actually do not affect the basic conformal algorithm too much, as presented in Algorithm \ref{alg_icp_mlm}. From here on, ``token" and ``word" will be used interchangeably. 

\begin{algorithm}
\SetAlgoLined
\KwResult{Returns the conformal prediction set $\Gamma^{\epsilon}$ containing candidate words for a masked token $d^* \in D_{\text{test}}$ and significance level $\epsilon$.}
\textbf{train} the model using $D_{\text{train}}$ to produce \\
\hspace{.25in} $\{\hat{y}_i^\text{mlm} : \ i \in \{1,\dots, n\}, d_i \in D_{\text{cal}}, \text{ and } d_i \text{ is masked}\}$\;
\For{$j$ in $\{i \in \{1,\dots, n\} : d_i \in D_{\text{cal}} \text{ and } d_i \text{ is masked}\}$}{
$u = y^{\text{mlm}}_j$; \hspace{.3in} {\footnotesize \# Recall $y^{\text{mlm}}_j$ is the true masked token} \\
$\alpha_j = 1 - \hat{y}^{\text{mlm}}_{j, u} $\;}
Re-index the nonconformity scores by $j \in \{1,\dots,\widetilde{k}\}$, where $\widetilde{k}$ is the number of sentences in $D_{\text{cal}}$ \;
\For{$u$ in $U$}{
$\alpha^*_{u} = 1 - \hat{y}^{\text{mlm}}_{*, u}$\;
$p_u = \frac{|\{j = 1,\dots,\widetilde{k} : \alpha_j \geq \alpha^*_u\}| + 1}{ \widetilde{k} + 1}$\;
\If{$p_u > \epsilon$}{
$u \in \Gamma^{\epsilon}$\;}
}
\Return{$\Gamma^{\epsilon}$}
\caption{\footnotesize ICP MLM}\label{alg_icp_mlm}
\label{alg_mlm}
\end{algorithm}

\begin{figure}[H]
    \centering
    \includegraphics[trim={58 0 0 30}, clip, scale=0.24]{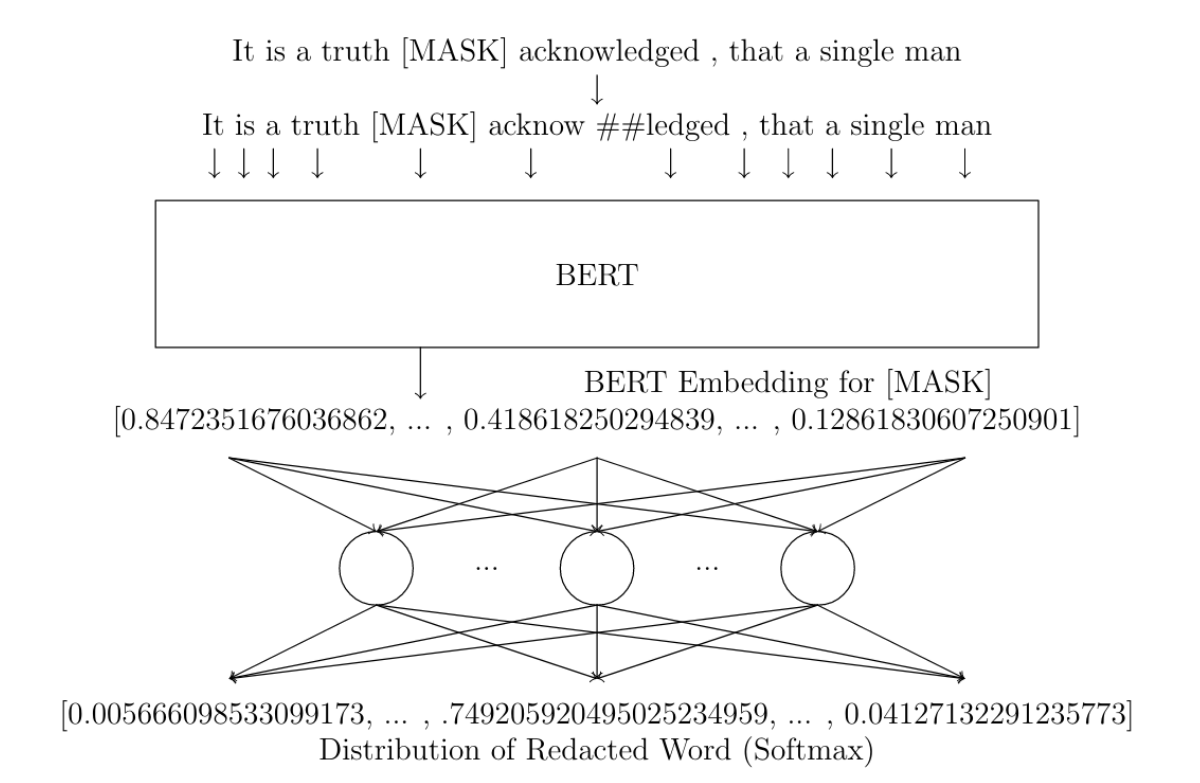}
    \caption{\footnotesize Illustration of the BERT MLM model. The top layer is the input sentence, which is then tokenized. A single token is then replaced with [MASK]. This tokenized sentence is then passed into BERT which outputs a softmax probability distribution corresponding to the masked token.}\label{fig:bert_mlm_schematic}
\end{figure}

For MLM, we construct a BERT-based conformal prediction algorithm similar to the BPS model for POS tagging described in the previous section. BERT was designed for the task of predicting a masked word. Our BERT model takes the context and position of a [MASK] token and returns a softmax distribution over the 30,522 candidate tokens, and then Algorithm \ref{alg_icp_mlm} is implemented to construct the conformal prediction set of candidate tokens for a given masked word of interest. Within Algorithm \ref{alg_icp_mlm}, $U$ denotes the set of all 30,522 unique tokens comprising the set of pre-defined BERT tokens. For the $j$-th masked token in $D$, $\hat{y}_j^\text{mlm} \in \mathbb{R}^{30,522}$ represents the softmax vector for the MLM model. In addition, $\hat{y}_{j, u}^\text{mlm}$ denotes the specific softmax value for any token $u \in U$.  A schematic of our BERT MLM is given in Figure \ref{fig:bert_mlm_schematic}.

\section{Empirical results}\label{results}

Using the Brown Corpus, we evaluate the conformal prediction sets produced by our three algorithms. The Brown Corpus contains 500 documents, with each word in these documents having a corresponding POS label. In total, there are just over 57,000 sentences and around 49,800 unique words. We consider each sentence in the corpus as a data instance and randomly allocate 80\% of these sentences for training, 10\% for calibration, and 10\% for testing. To account for sampling variability, the random allocation of the data into training, calibration, and testing sets is repeated 5 times, and all metrics are evaluated on and averaged over the 5 test sets. 

For all POS tags (and combination of POS tags) we remove the hyphenated portion (if any). This includes  headline (-HL), title (-TL), and emphasis (-NC) hyphenations, as well as foreign word prefix (FW-). If a word has a POS listed as a combination of multiple POS, the specific multiple POS combination is added as a new unique POS to our label set. After these preprocessing steps there remain $q = 190$ unique POS tags in the label set.

\subsection{Performance metrics}

We consider a variety of metrics that evaluate both the ``forced'' point-predictions and the conformal prediction sets. The metrics we consider are adopted from the criteria considered in \cite{Maltoudoglou2020}. Let $n_{\text{test}} := |D_{\text{test}}|$, and assume a fixed $\epsilon \in (0,1)$. For ease of notation, let $\hat{y}_i$ denote a prediction for some label $y_i$, for some example indexed by $i$. The metrics are defined as follows.

Classification accuracy ($CA$) is taken simply to be the proportion of correct predictions:
\[
CA = \frac{1}{n_{\text{test}}}\sum_{i=1}^{n_{\text{test}}} I[\hat{y}_i = y_i].
\]
Average credibility ($\overline{Cred}$) is the average minimum significance level required such that our prediction sets are nonempty:
\[
\overline{Cred} = \frac{1}{n_{\text{test}}}\sum_{i=1}^{n_{\text{test}}} \inf \{1-\epsilon \,:\, \qty|\Gamma^{\epsilon}_i| \geq 1 \}.
\]
A high value for $\overline{Cred}$ is an indication that the model has little confidence that {\em any} of the considered labels are appropriate for the test examples. The $OP$ criterion (for {\em observed perceptiveness}) is the average of all test p-values for correct classifications:
\[
OP = \frac{1}{n_{\text{test}}}\sum^{n_{\text{test}}}_{i=1} p_{y_i}.
\]
Conversely, the $OF$ criterion (for {\em observed fuzziness}) is the average of all test p-values for incorrect classifications:
\[
OF = \frac{1}{n_{\text{test}}}\sum^{n_{\text{test}}}_{i=1} \sum_{y\ne y_i} p_{y}.
\]
Average empirical coverage ($\overline{Coverage}$) is the proportion of prediction sets that contain the true value:
\[
\overline{Coverage} = \frac{1}{n_{\text{test}}}\sum_{i=1}^{n_{\text{test}}} I[y_i \in \Gamma_i^\epsilon].
\]
Proportion of indecisive sets ($PIS$) is the proportion of sets (for a fixed $\epsilon$) that contain more than one label:
\[
PIS = \frac{1}{n_{\text{test}}}\sum_{i=1}^{n_{\text{test}}} I[|\Gamma_i^\epsilon| > 1].
\]
The average confidence of decisive sets ($ACDS$) is the proportion of confidence sets of size 1 that contain the true label:
\[
ACDS = \frac{\sum_{i=1}^{n_{\text{test}}} I[|\Gamma_i^\epsilon| = 1 , y_{i} \in \Gamma_i^\epsilon]}{\sum_{i=1}^{n_{\text{test}}} I[|\Gamma_i^\epsilon| = 1]}.
\]
Lastly, the $N_{\epsilon}$ criterion is the mean size of prediction sets at level of significance $1 - \epsilon$:
\[
N_{\epsilon} = \frac{1}{n_{\text{test}}}\sum_{i=1}^{n_{\text{test}}}|\Gamma^\epsilon_i|.
\]

\subsection{POS prediction results} 

Figure \ref{table:pos_sets} and \ref{table:pos_forced} present the results for both POS models.  Note that the metrics in Figure \ref{table:pos_forced} require a forced point-prediction, which we take to be the label that maximizes the softmax vector that is returned by either the BPS model or the BiLSTM model. 

\begin{figure}[H]
\begin{tabular}{|c|c||c|c|c|}
\hline
                        & Proposed Conf.                             & 99.9\% & 99\%   & 95\%                        \\ \hline\hline
\multirow{4}{*}{BiLSTM} & \small{$\overline{Coverage}$} & 0.9989  & 0.9903 & 0.9502                      \\ \cline{2-5} 
                        & $ACDS$                                      & 0.9996  & 0.9939 & 0.9631                      \\ \cline{2-5} 
                        & $PIS$                                        & 0.5424  & 0.1566 & 0.0024                      \\ \cline{2-5} 
                        & $N_{\epsilon}$                                   & 3.4336  & 1.2732 & 0.9889                      \\ \hline\hline
\multirow{4}{*}{BPS}    & \small{$\overline{Coverage}$} & 0.9990  & 0.9897 & 0.9499                      \\ \cline{2-5} 
                        & $ACDS$                                       & 0.9992  & 0.9909 & NA                          \\ \cline{2-5} 
                        & $PIS$                                        & 0.3577  & 0.0334 & 0.0000                      \\ \cline{2-5} 
                        & $N_{\epsilon}$                                   & 2.6260  & 1.0378 & \multicolumn{1}{c|}{0.9570} \\ \cline{5-5} 
\hline
\end{tabular}\caption{\footnotesize Set-value prediction criterion results for POS prediction}\label{table:pos_sets}
\end{figure}

It is observed in Figure \ref{table:pos_sets} that for the 99\% nominal confidence level, both models produce sets that average around 1-2 POS per set. This illustrates that the conformal prediction algorithm produces efficient sets at high confidence levels, and also suggests that the softmax probability vectors from the underlying neural nets are highly concentrated on 1-2 POS labels. Moreover, the $\overline{Coverage}$ and $ACDS$ values in Figure \ref{table:pos_sets} demonstrate that these conformal prediction sets achieve their nominal coverage.  Excessively small values for $PIS$ with $N_{\epsilon} \approx 1$ at the 95\% confidence level indicate a high proportion of conformal prediction sets containing zero or one POS label.

\begin{figure}[H]
\begin{tabular}{|c|c|c|c|c|}
\hline
Model  & $CA$     & \small $\overline{Cred.}$   & $OP$     & $OF$  \\
\hline
BiLSTM & 0.9536 & 0.5055 & 0.5012 & 0.0493 \\
\hline
BPS    &  0.9793      &   0.5020     &   0.5008     &    0.0126    \\
\hline
\end{tabular}\caption{\footnotesize Forced-value prediction criterion results for POS prediction}\label{table:pos_forced}
\end{figure}

\begin{figure}[H]
\centering
\includegraphics[trim={0 20 0 20}, clip, scale=.425]{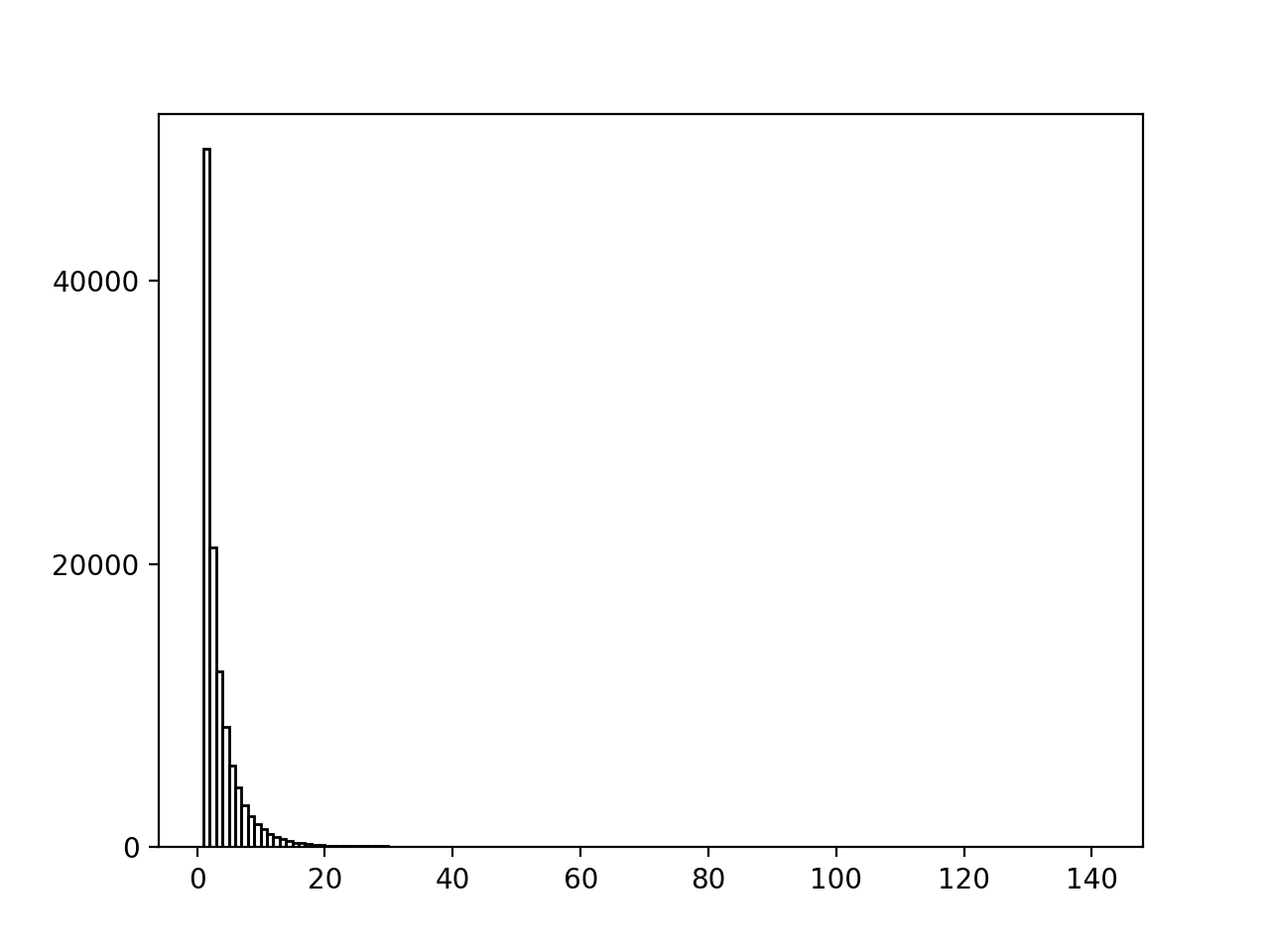}\\
\includegraphics[trim={0 20 0 20}, clip, scale=.425]{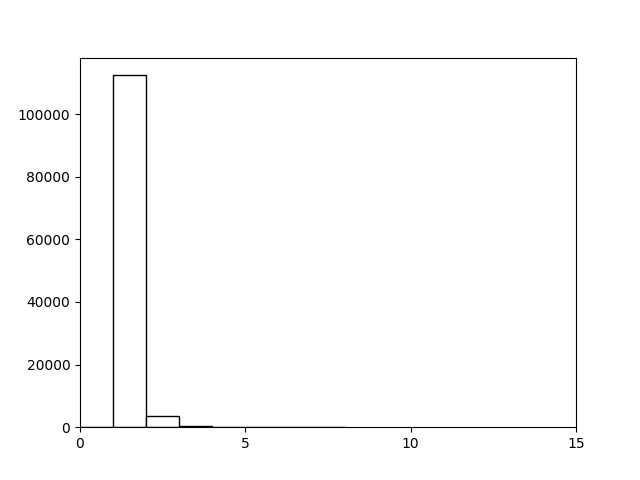}\\
Set size
\caption{\footnotesize Histograms of conformal prediction set sizes for POS prediction at the 99\% confidence level for BiLSTM (top) and BPS (bottom).}
\label{fig:pos_set_sizes}
\end{figure}

To offer further insight, Figure \ref{fig:pos_set_sizes} displays histograms of the set sizes for both models at the 99\% confidence level. The majority of the sets are of size one, which accounts for the height of the leftmost bins. 
However, the sizes of the sets vary greatly for different levels of nominal confidence, and so the uncertainty quantification afforded by the conformal prediction sets has utility.  In particular, the models we constructed are able to provide 99.9\% confidence for 3-4 POS labels, on average, for a given word.  Such a quantified guarantee about the uncertainty in a prediction is not possible to provide from neural network architectures alone.

\begin{figure}[H]
\centering
\rotatebox{90}{\hspace{.15in}\footnotesize{Empirical Confidence Level}}
\includegraphics[trim={0 20 0 20}, clip, scale=.425]{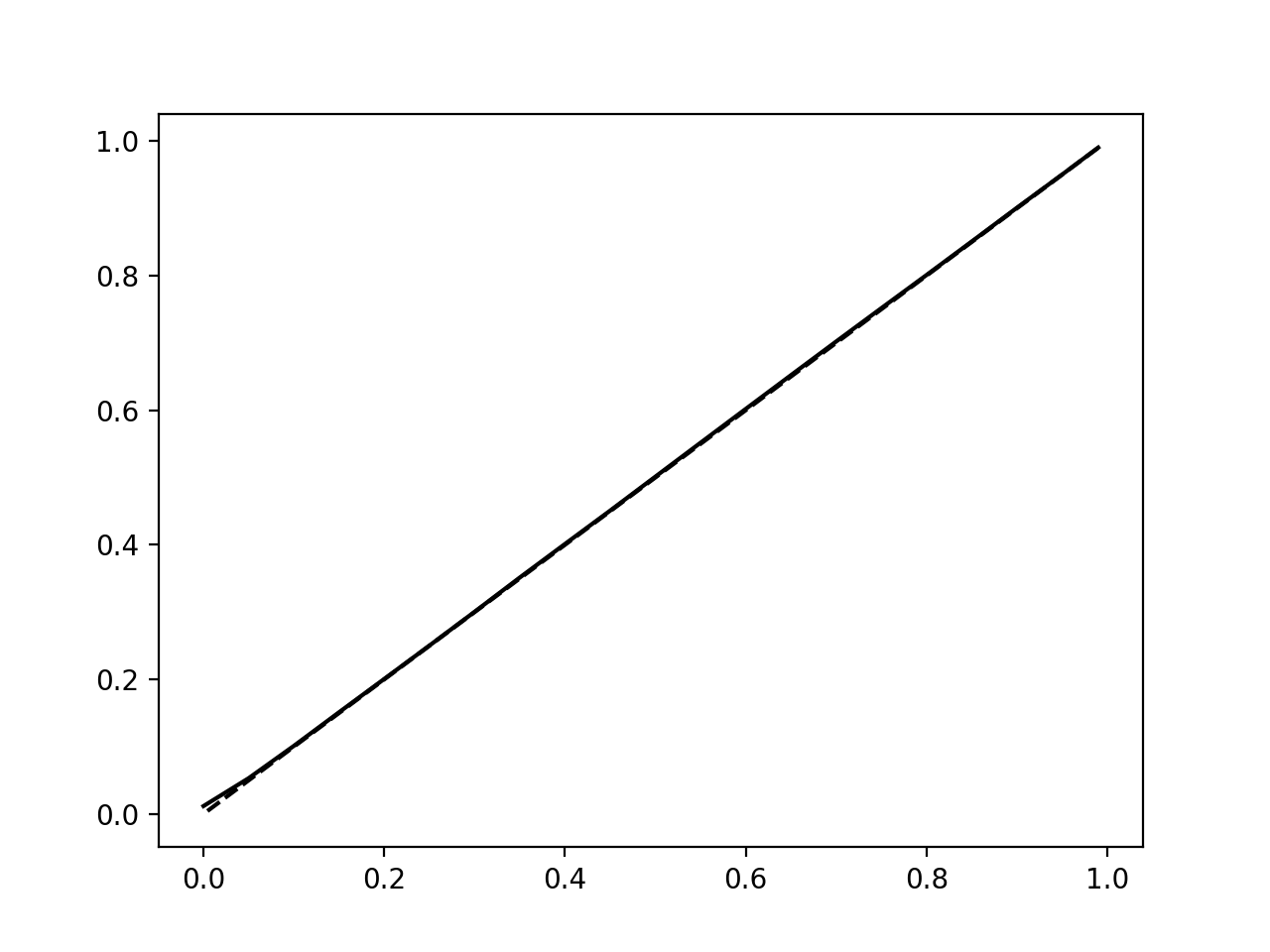}\\
\rotatebox{90}{\hspace{.15in}\footnotesize{Empirical Confidence Level}}
\includegraphics[trim={0 20 0 20}, clip, scale=.425]{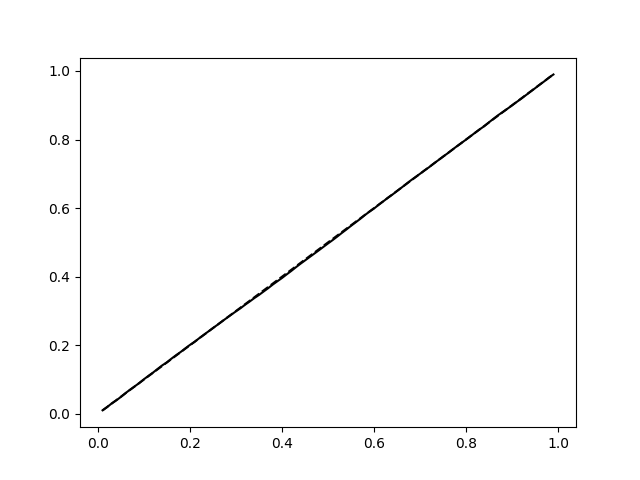}\\
\footnotesize{Nominal confidence level}
\caption{\footnotesize $\overline{Coverage}$ of conformal prediction sets for POS prediction for BiLSTM (top) and BPS (bottom). For reference, the dashed line is a 45 degree line.}
\label{fig:pos_coverage}
\end{figure}

\begin{figure}[H]
\centering
\includegraphics[trim={0 20 0 20}, clip, scale=.39]{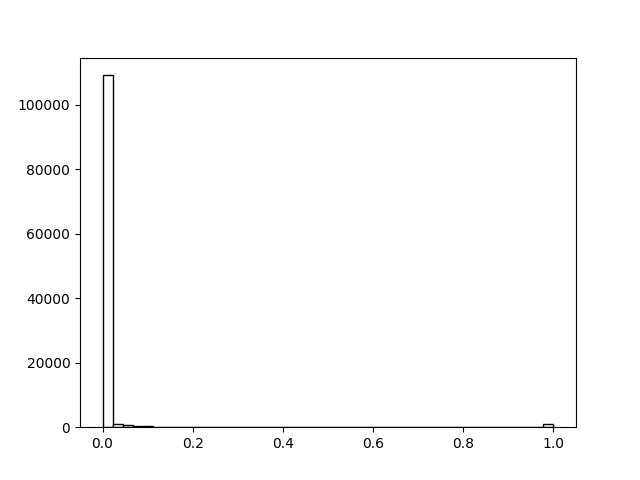}\\
\includegraphics[trim={0 20 0 20}, clip, scale=.39]{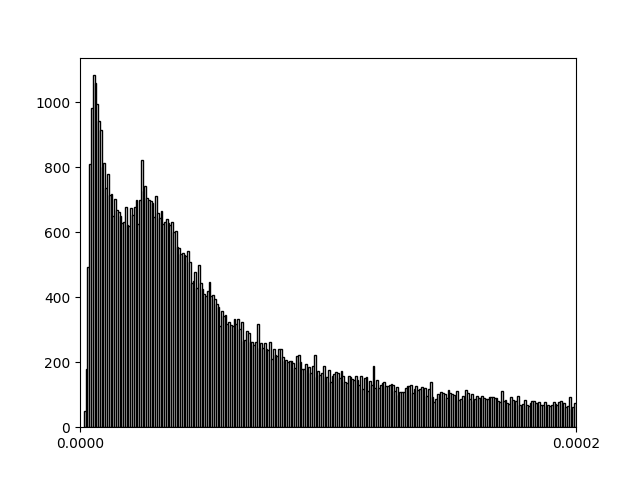}\\
Nonconformity score
\caption{\footnotesize \footnotesize Histogram of nonconformity scores for the calibration sets for the BPS model. The histogram on the bottom plot only includes scores less than 0.0002 to better illustrate how these scores are distributed near zero.}\label{fig:pos_scores_bps}
\end{figure}

\begin{figure}[H]
\centering
\includegraphics[trim={0 20 0 20}, clip, scale=.39]{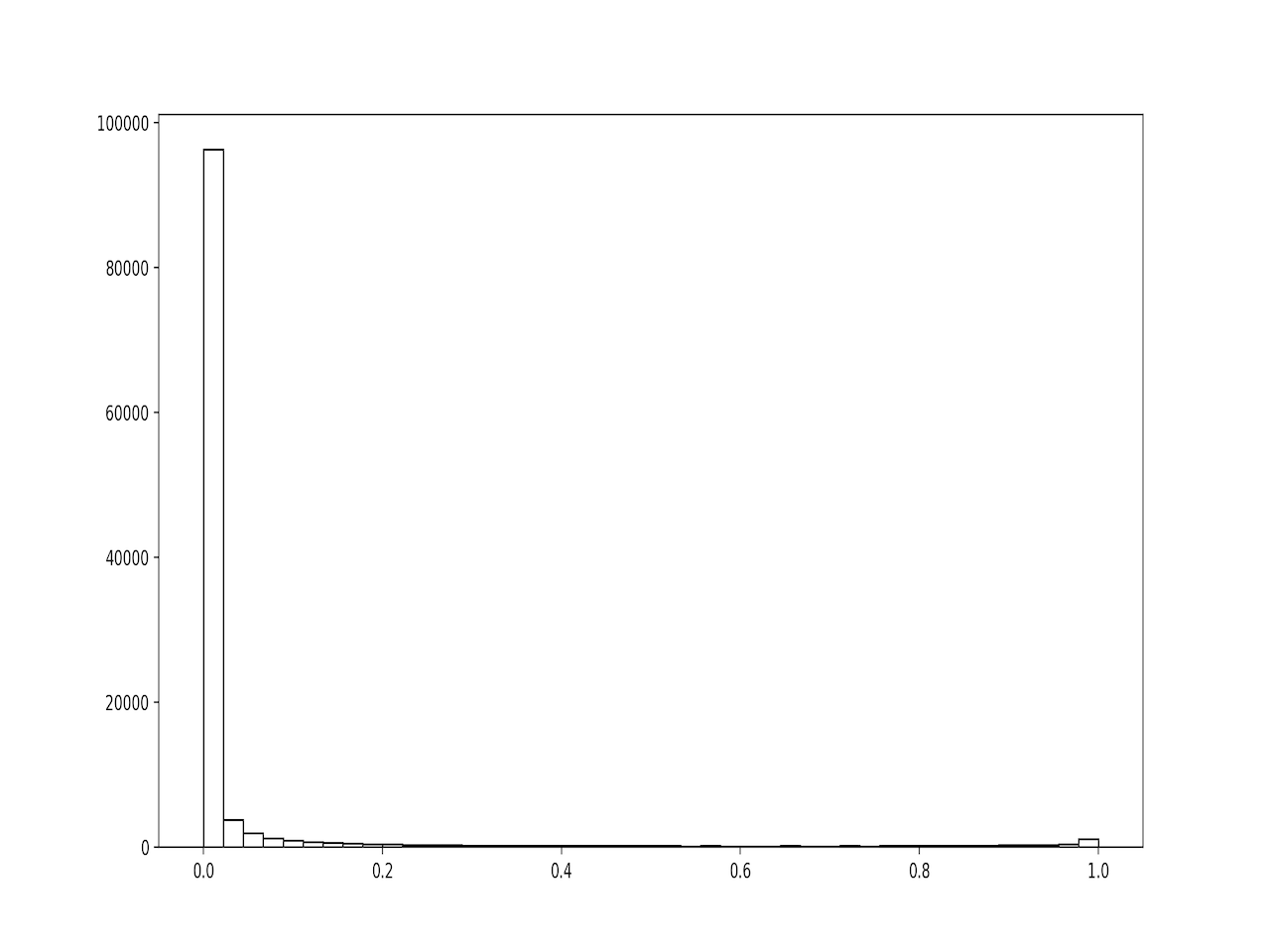}
\includegraphics[trim={0 20 0 20}, clip, scale=.39]{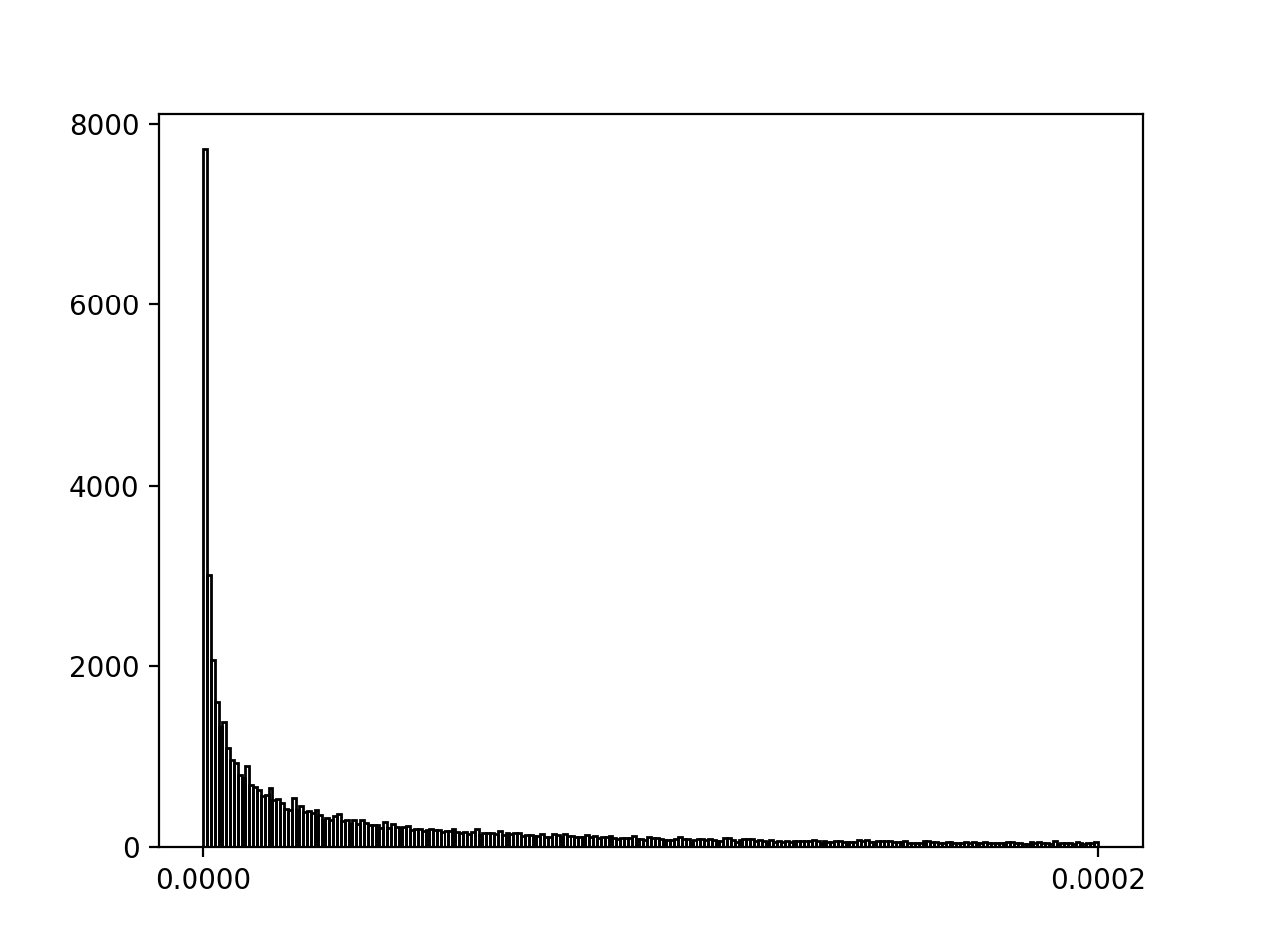}\\
Nonconformity score
\caption{\footnotesize \footnotesize Histogram of nonconformity scores for the calibration sets for the BiLSTM POS model. The histogram on the bottom plot only includes scores less than 0.0002 to better illustrate how these scores are distributed near zero.}\label{fig:pos_scores_BiLSTM}
\end{figure}

To demonstrate the validity for values of $\overline{Coverage}$ at more levels than the 99.9\%, 99\%, and 95\% levels displayed in Figure \ref{table:pos_sets}, Figure \ref{fig:pos_coverage} plots the average empirical coverage of the conformal prediction sets against their nominal levels for levels of significance ranging from 0 to 1.

Next, Figure \ref{table:pos_forced} provides an assessment of the forced point-predictions of the underlying BPS and BiLSTM models. Being the state-of-the-art, it is found that the BPS model is marginally more accurate with respect to $CA$. However, both models perform relatively similar with regard to the other metrics in Figure \ref{table:pos_forced}. The difference in values between $OP$ and $OF$ indicate that the models are able to discriminate the correct POS label from the incorrect labels, on average.

Lastly, for further assessment of the conformal prediction algorithm, we present histograms of the nonconformity scores for the calibration sets in Figures \ref{fig:pos_scores_BiLSTM} and \ref{fig:pos_scores_bps}.

\subsection{MLM results}

For the MLM task, we mask a randomly chosen single word in each sentence in the Brown Corpus.  Sentences are tokenized according to the ``WordPiece" embeddings used by BERT, then truncated to a length of 128 to feed into the model.  Further, we include fewer examples in the calibration set for the MLM task than in the previous section for the POS task due to the larger computational cost entailed by the much larger label set for MLM (i.e., all words in a vocabulary of around 30,000 words). Specifically, the calibration set contains around 1,300 sentences, and the testing set is also reduced to 1,000 sentences.  To account for sampling variability in the random allocation of the data into training, calibration, and testing sets, we still repeat the process 5 times and report our results as averages of these 5 Monte Carlo iterations.

\begin{figure}[H]
\begin{tabular}{|l|l|l|}
\hline
Confidence Level & $\overline{Coverage}$ & $N_{\epsilon}$\\
\hline
95\%             & .948  & 176.77\\
\hline
90\%             & .898 & 43.62 \\
\hline
80\%             & .794 & 6.96 \\
\hline
75\%             & .739 & 3.65\\
\hline
\end{tabular}\caption{\footnotesize Set-value prediction criterion results for MLM}\label{table:mlm_sets}
\end{figure}

\begin{figure}[H]
\begin{tabular}{|c|c|c|c|c|}
\hline
Model   & CA   & $\overline{Cred}$ & OP  & OF \\
\hline
BERT MLM & 0.542 & 0.609 & .491 & .122\\
\hline
\end{tabular}\caption{\footnotesize Forced-value prediction criterion results for MLM}\label{table:mlm_forced}
\end{figure}

\begin{figure}[H]
\centering
\includegraphics[trim={0 20 0 20}, clip, scale=.45]{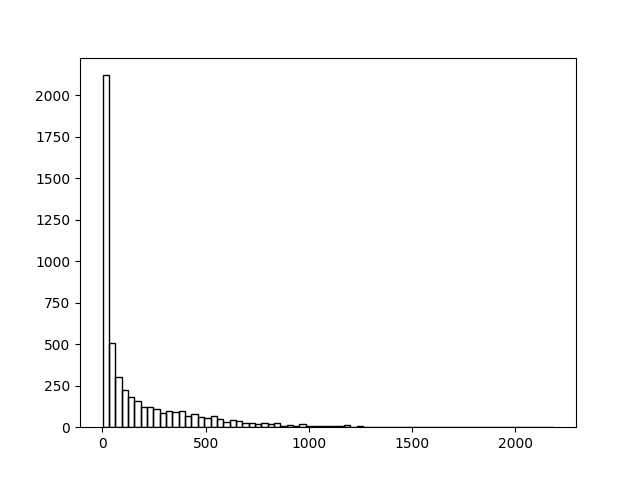} \\
Set size
\caption{\footnotesize Histograms of conformal prediction set sizes for MLM at the 95\% confidence level.}
\label{fig:mlm_set_sizes}
\end{figure}

Unlike for POS prediction in the previous section, for MLM it is found that higher levels of confidence lead to prediction sets that are too large to be useful (see Figure \ref{table:mlm_sets}).  In particular, to guarantee that the true masked word is not omitted from the prediction set for more than 5\% of test sentences (i.e., at the 95\% level), the average conformal prediction set size is reported to be approximately 177 candidate tokens. Nonetheless, sacrificing some confidence quickly leads to smaller sets, down to 3-4 words on average at the 75\% level. The histogram of the conformal prediction set sizes for all test examples is shown in Figure \ref{fig:mlm_set_sizes}.

\begin{figure}[H]
\centering
\rotatebox{90}{\hspace{.25in}\footnotesize{Empirical confidence level}}
\includegraphics[trim={0 20 0 20}, clip, scale=.45]{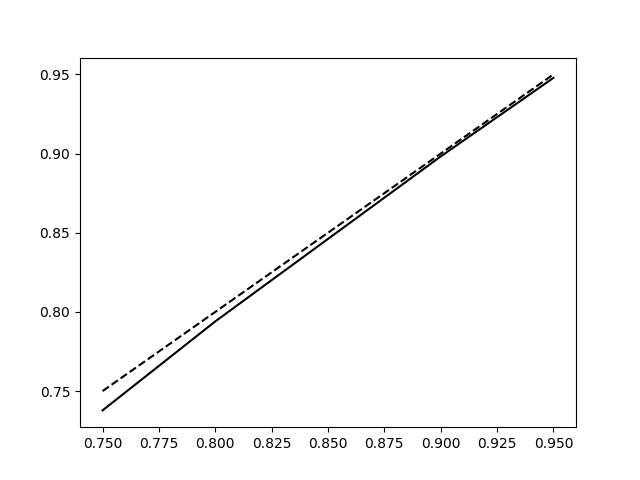}\\
Nominal confidence level
\caption{\footnotesize $\overline{Coverage}$ of conformal prediction sets for MLM. For reference, the dashed line is a 45 degree line.}\label{fig:mlm_coverage}
\end{figure}

\begin{figure}[H]
\centering
\includegraphics[trim={0 20 0 20}, clip, scale=.45]{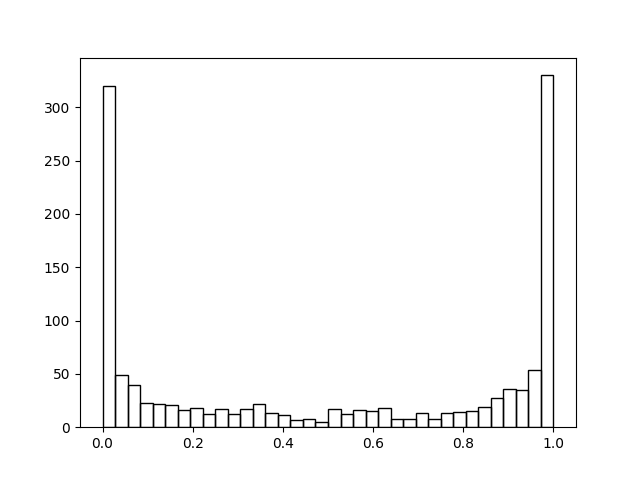}\\
Nonconformity score
\caption{\footnotesize Histogram of nonconformity scores for the calibration sets for MLM.}\label{fig:mlm_scores}
\end{figure}

Additionally, the conformal prediction sets do achieve their nominal coverage at all levels displayed in Figure \ref{table:mlm_sets}. To infer the validity for all values of $\overline{Coverage}$ from 75\% to 95\%, Figure \ref{fig:mlm_coverage} plots the average empirical coverage of the conformal prediction sets against their nominal levels of significance in this range.

Lastly, we provide the forced point-prediction metrics in Figure \ref{table:mlm_forced}, and we present a histogram of the nonconformity scores for the calibration sets in Figure \ref{fig:mlm_scores}. The bimodal nature of the histogram is due to the underlying BERT model making overly discriminative predictions (i.e., the softmax vectors $\hat{y}_j^\text{mlm}$ being close to a one-hot vector), even when these predictions are sometimes very wrong, leading to either very high or very low nonconformity scores and not much in-between.

\section{Illustrative real example}\label{real_data}

An application of our conformal prediction sets for MLM could come in the form of a post-hoc analysis tool for speech recognition software. The following example comes from a voice transcription of a 2009 TED Talk given by Michelle Obama, part of the greater TED-LIUM3 audio transcription corpus \citep{Hernandez2018}. However, not all words were able to be detected by the automated speech recognition (ASR) system, and are instead labeled with the token $<$UNK$>$ to take the place of the unknown word. Ideally, our model would be able to fill in these unknown words with set-valued predictions for any desired confidence level. To compare with other voice-to-text systems, we also analyzed the YouTube closed-captioning for this TED Talk video, which appeared to be more accurate than the ASR. Below are 3 example sentences from the talk, with the italicized text representing the YouTube closed-captioning transcriptions, and the non-italicized text representing the ASR system transcriptions. The correct words, along with conformal prediction sets at the 75\% confidence level (i.e., $\epsilon = 0.25$), are presented next.

\bigskip
{\noindent\bf Example 1}.
\begin{quote}
\textit{\dots to go with him to a community meeting. But when we met, Barack was a community organizer.}
\end{quote}
\begin{quote}
\dots to go with him to a community $<$UNK$>$. But when we met, Barack was a community organizer.
\end{quote}
\noindent $\Gamma^{0.25} = $ [`college', `center', `event', `conference', `meeting', `dinner', `gathering']\\
{\noindent\bf Correct word:} `meeting'
\hfill $\blacksquare$

\bigskip
{\noindent\bf Example 2}.
\begin{quote}
\textit{And he urged the people in that meeting, in that community, to devote themselves to closing the gap between those two ideas, to work together to try to make the world as it is and the world as it should be, one and the same.}
\end{quote}
\begin{quote}
And he urged the people in that meeting in that community to devote themselves to closing the gap between those two ideas, to work together to try to make the world as it is and the world as it should $<$UNK$>$ one and the same. 
\end{quote}
\noindent $\Gamma^{0.25} = $ [`,', `be', `seem']\\
{\noindent\bf Correct word:} `be'
\hfill $\blacksquare$

\bigskip
{\noindent\bf Example 3}.
\begin{quote}
\textit{And they opened many new doors for millions of female doctors and nurses and artists and authors, all of whom have followed them. And by getting a good education you too can control your own destiny.}
\end{quote}
\begin{quote}
And they opened many new doors for millions of female doctors and nurses and artists and authors all of whom have $<$UNK$>$ $<$UNK$>$. And by getting a good education you too can control your own destiny.  
\end{quote}
\noindent $\Gamma_{1}^{0.25} = $ [`been', `become', `loved']\\
\noindent $\Gamma_{2}^{0.25} = $ [`children', `died', `success', `experience', `careers']\\
{\noindent\bf Correct word:} `followed', `them'
\hfill $\blacksquare$

\bigskip
At the 75\% confidence level, the conformal prediction sets included the correct word in the first two examples. However, our MLM was not trained on any sentence with two consecutive masked words, thus it fails to include the correct words in the third example. That being so, if we pass this sentence through the model twice, each time with only one masked word, we see the more accurate results: \\

{\noindent\bf Correct word:} `followed' \\
\noindent $\Gamma_{2}^{0.25} = $ [`joined', `followed', `loved', `taught', `inspired', `influenced']\\
    
{\noindent\bf Correct word:} `them' \\
\noindent $\Gamma_{2}^{0.25} = $ [`you', `me', `them', `through', `suit']\\

This suggests that the BERT model heavily depends on directly adjacent words to predict the token for a masked word in a sentence.

\section{Concluding remarks}\label{conclusion}  

We found that BERT-based conformal prediction sets were extremely effective in predicting both POS and masked words, which is unsurprising seeing as BERT is the dominant model for many NLP tasks at the moment.  The complexity of models like BERT or BiLSTM was necessary, as our previous attempts using simpler nonconformity functions were not able to produce as efficient confidence sets. In the future, we may explore different nonconformity scores to get the BERT MLM prediction intervals even smaller. Initial tests show promising results, but these are more computationally intensive than the methods described in our results section.

\bibliographystyle{plain}
\bibliography{citations.bib}

\end{document}